# Cross-Lingual Consistency: A Novel Inference Framework for Advancing Reasoning in Large Language Models

*Zhiwei Yu[1,2]\*, Tuo Li[1,2], Changhong Wang[1,2], Hui Chen[3], Lang Zhou[1,2]*

[1]*Shandong Inspur Artificial Intelligence Research Institute Co., Ltd. Jinan 100089, Shandong, China*

[2]*Shandong Yunhai Guochuang Cloud Computing Equipment Industry Innovation Center Co., Ltd., Jinan 100089, Shandong, China*

[3]*BNRist,Tsinghua University, Haidian District, Beijing, 100084, China*

yuzhiwei01@inspur.com, jichenhui2012@gmail.com

## Abstract

Chain-of-thought (CoT) has emerged as a critical mechanism for enhancing reasoning capabilities in large language models (LLMs), with self-consistency demonstrating notable promise in boosting performance. However, inherent linguistic biases in multilingual training corpora frequently cause semantic drift and logical inconsistencies, especially in sub-10B parameter LLMs handling complex inference tasks. To overcome these constraints, we propose the cross-lingual consistency (CLC) framework, an innovative inference paradigm that integrates multilingual reasoning paths through majority voting to elevate LLMs' reasoning capabilities. Empirical evaluations on the CMATH dataset reveal CLC's superiority over the conventional self-consistency method, delivering 9.5%, 6.5%, and 6.0% absolute accuracy gains for DeepSeek-Math-7B-Instruct, Qwen2.5-Math-7B-Instruct, and Gemma2-9B-Instruct respectively. Expanding CLC's linguistic scope to 11 diverse languages implies two synergistic benefits: 1) neutralizing linguistic biases in multilingual training corpora through multilingual ensemble voting, 2) escaping monolingual reasoning traps by exploring the broader multilingual solution space. This dual benefits empirically enables more globally optimal reasoning paths compared to monolingual self-consistency baselines, as evidenced by the 4.1%-18.5% accuracy gains using Gemma2-9B-Instruct on the MGSM dataset.

# 1 Introduction

The emergence of large language models (LLMs) has revolutionized natural language processing (NLP) capabilities across diverse applications (Minaee et al., 2024; Naveed et al., 2024). A critical component maximizing LLM efficacy lies in prompt engineering(Brown et al., 2020; Radford et al., 2019), which involves designing and refining the input prompts to elicit more accurate, relevant, and focused outputs. Among recent breakthroughs, chain-of-thought (CoT) prompting (Wei et al., 2023) has emerged as a transformative strategy.

CoT prompting has proven particularly valuable in reasoning tasks that inherently demand multi-step logical inference to derive accurate conclusions. Traditional large language models (LLMs) demonstrate the capability to generate immediate responses, yet fundamentally lack the capacity to externalize their internal reasoning pathways or justify conclusions through explicit logical progression. The CoT methodology bridges this capability gap by structurally guiding models to articulate intermediate reasoning steps, thereby significantly enhancing performance on complex multi-step tasks requiring systematic analysis. Furthermore, the self-consistency technique (Wang X. et al., 2023) employs sampling strategies such as beam search and nucleus sampling (Holtzman et al., 2020) instead of greedy decoding, which has been empirically shown to amplify CoT's effectiveness in domains requiring rigorous deductive processes—including mathematical reasoning, symbolic logic operations, and causal inference

(Chen et al., 2023; P. Wang et al., 2023; Shao et al., 2024).

Despite these advancements, several challenges remain in the reasoning domain. Firstly, although CoT prompting has demonstrated substantial improvements in multi-step reasoning tasks, models exhibit constrained generalization capacity when handling novel complexity scenarios. LLMs still face difficulties when tasked with reasoning about unfamiliar domains or situations not well-represented in the training data (Stechly et al., 2025). Secondly, while the self-consistency methodology enhances models' performance by sampling the reasoning paths over breadth, issues of reasoning incorrectness persist (Chen et al., 2023; P. Wang et al., 2023), particularly when most of the sampled reasoning paths trapped in local optima. Notably, small-scale LLMs are particularly susceptible to the inherent linguistic biases present in multilingual training corpora. As a result, they often exhibit semantic drift and logical inconsistencies, leading to poor performance on low-resource languages (Lample & Conneau, 2019; Xu et al., 2024; Bajpai & Chakraborty, 2025).This correlates with evidence that sub-10B parameters LLMs have limited capacity and struggle with sampling deep reasoning paths (Kaplan et al., 2020; Wei et al., 2023).

This paper aims to address these limitations by proposing a novel inference framework, named cross-lingual consistency (CLC), to improve the reasoning capabilities of LLMs. Specifically, we substitute the traditional CoT prompts in monolingual self-consistency with multilingual CoT ensembles, where parallel reasoning paths across diverse languages undergo majority voting. Empirical results demonstrate that CLC achieves 4.1%–18.5% absolute accuracy gains on the MGSM benchmark compared to monolingual self-consistency baselines, with particularly pronounced improvements in low-resource language scenarios. Through the CLC inference framework, we hope to push the boundaries of LLMs' reasoning capabilities and pave the way for more accurate, reliable, and interpretable AI systems.

## 2 Method

We propose an inference framework named CLC, as shown in Figure 1. Given the original textual sequence $x$ representing the problem to be reasoned, the proposed framework consists of three steps as follows: (1) Initially, the problem $x$ is translated and expanded into multiple language versions. For instance, if the input problem is in Chinese, it is translated into English and French. The selection of target languages primarily depends on the language repertoire of the specified LLM. It is noteworthy that the prompt templates employed in this study are also translated into corresponding languages to ensure compatibility with the input problems. (2) Subsequently, the LLM is prompted to generate multiple candidate answers for the problem in each selected language, thereby yielding a diverse set of potential solutions. (3) The generated candidate answers are then aggregated and statistically analyzed to determine the final answer. For deterministic tasks, where the problem has a unique solution (e.g., mathematical problems), majority vote is used as the decision rule. Specifically, all generated candidate answers are consolidated into a single set. The frequency of each candidate answer within this set is tallied, and the answer with the highest frequency is selected as the final solution. For non-deterministic tasks, where the problem admits multiple valid solutions (e.g., text summarization), a semantic clustering approach or an LLM can be employed to determine the final answer. Taking semantic clustering as an example, the process involves clustering the semantic vectors of the candidate answers and selecting the answer that is semantically closest to the centroid of the most populous cluster as the final solution.

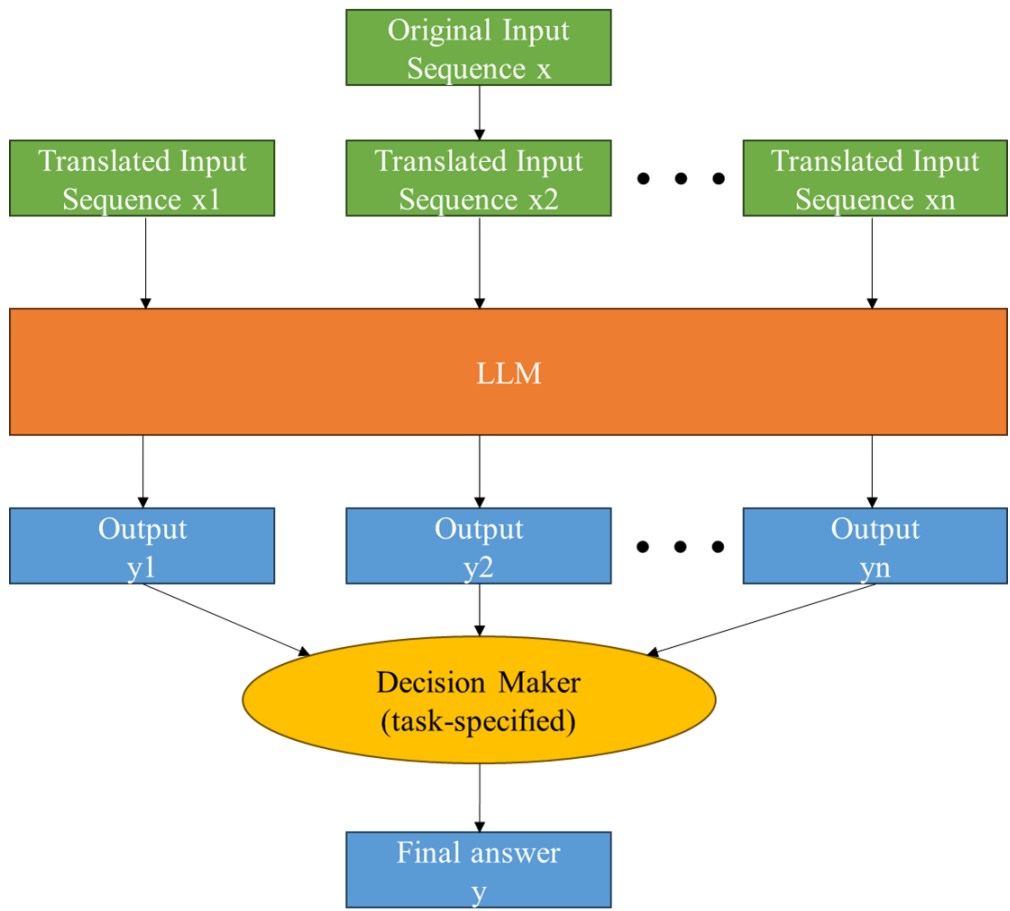

**Figure 1**. Overview of the cross-lingual consistency framework. The question *x* that requires reasoning is firstly expanded into different language versions through translation. For example, if the original input question is in Chinese, it will be translated into English and French. Secondly, the LLM explores diverse reasoning paths and generate candidate answers in diverse languages. Finally, the Decision maker decides the final answer.

**3 Experiment**

### 3.1 Experiment Setup

**Datasets.** We evaluated CLC inference framework on two math datasets, including CMATH and MGSM. The original CMATH dataset is in Chinese, so we used Google Translate to translate the questions of CMATH in English and evaluated the bilingual-consistency inference framework. We later expanded bilingual-consistency to multilingual-consistency framework on the MGSM dataset with 11 languages, including English, Spanish, French, German, Russian, Chinese, Japanese, Thai, Swahili, Bengali, and Telugu.

**Models**. Since small-scale LLMs face challenges in exploring reasoning paths at sufficient depth and are sensitive to linguistic biases in multilingual training corpora, it is likely to boost performance of small-scale LLMs using CLC which samples the reasoning paths over breadth and makes full use of training data in different languages. Therefore, this paper mainly focuses on small-scale open-source models, including Qwen2.5-Math-7B-Instruct, Gemma2-9B-Instruct, and DeepSeek-Math-7B-Instruct. In order to accelerate inference, vLLM framework was deployed and model weights were quantized to 16-bit precision (using float16 for DeepSeek-Math-7B-Instruct and bfloat16 for Qwen2.5-Math-7B-Instruct and Gemma2-9B-Instruct).

**Prompts**. Zero-shot CoT prompts are utilized to instruct LLMs to reason step by step. The Prompts for DeepSeek-Math-7B-Instruct are the same as those officially provided by DeepSeek, as shown in Table 1. Qwen2.5-Math-7B-Instruct and Gemma2-9B-Instruct also use the same prompts as DeepSeek-Math-7B-Instruct, since there is no officially recommended CoT prompts for them.

| Model | Prompts |
|---|---|
| DeepSeek-Math-7B-Instruct | **(1) English Prompt:**<br>Question\n Please reason step by step, and put your final answer within \\boxed{}.<br>**(2) Chinese Prompt:**<br>Question\n 请通过逐步推理来解答问题，并把最终答案放置于\\boxed{}中。 |
| Gemma2-9B-Instruct | **(1) English Prompt:**<br>Question\n Please reason step by step, and put your final answer within \\boxed{}.<br>**(2) Chinese Prompt:**<br>Question\n 请通过逐步推理来解答问题，并把最终答案放置于\\boxed{}中。<br>**(3) Spanish Prompt:**<br>Question\n Razone paso a paso y coloque su respuesta final dentro de \\boxed{}.<br>**(4) French Prompt:**<br>Question\n Veuillez raisonner étape par étape et mettre votre réponse finale dans \\boxed{}.<br>**(5) German Prompt:**<br>Question\n Bitte begründen Sie Schritt für Schritt und geben Sie Ihre endgültige Antwort in \\boxed{} ein.<br>**(6) Russian Prompt:**<br>Question\n Пожалуйста, рассуждайте шаг за шагом и поместите окончательный ответ в \\boxed{}.<br>**(7) Japanese Prompt:**<br>Question\n 段階的に推論して、最終的な答えを \\boxed{} 内に入力してください。<br>**(8) Thai Prompt:**<br>Question\n โปรดให้เหตุผลทีละขั้นตอน และใส่คำตอบสุดท้ายของคุณไว้ใน \\boxed{}<br>**(9) Swahili Prompt:**<br>Question\n Tafadhali sababu hatua kwa hatua, na uweke jibu lako la mwisho ndani ya \\boxed{}.<br>**(10) Bengali Prompt:**<br>Question\n অনুগ্রহ করে ধাপে ধাপে যুক্তি দিন এবং আপনার চূড়ান্ত উত্তরটি \\boxed{}-এর মধ্যে রাখুন।<br>**(11) Telugu Prompt:**<br>Question\n దయచేసి దశల వారీగా వాదించండి మరియు మీ తుది సమాధానాన్ని \\ బాక్స్డ్{}లో ఉంచండి. |

**Table 1**. Zero-shot CoT prompts for DeepSeek-Math-7B-Instruct and Gemma2-9B-Instruct.

### 3.2 Results

### 3.2.1 Bilingual-Consistency

The effectiveness of CLC was initially verified in the case of two languages (Chinese and English), i.e., bilingual-consistency. Figure 2 shows the performance of DeepSeek-Math-7B-Instruct on CMATH using bilingual-consistency, traditional self-consistency and sampling without consistency. Sampling Chinese reasoning paths without consistency shows an accuracy range of 76.3% to 78.5% over 10 epochs, while sampling English reasoning paths without consistency shows an accuracy range of 50.0% to 54.7% over 10 epochs. Hence, compared with English reasoning, DeepSeek-Math-7B-Instruct is much better at Chinese reasoning on CMATH. The sampling answers over 10 epochs were then integrated to obtain the traditional self-consistency results using majority voting. The accuracy curves of self-consistency in Chinese and English exhibit a similar trend: a rapid initial rise during the early training epochs, followed by a gradual deceleration in improvement. As demonstrated in previous research (Wang X. et al., 2023), self-consistency does boost the performance of CoT prompts, leading to 85.8% and 77.3% accuracy at the 10th epoch in Chinese and English, respectively. Finally, bilingual-consistency results were obtained by integrating the results of the 10 Chinese reasoning epochs and 10 English reasoning epochs via majority voting, as shown in Figure 2. Specifically, the bilingual-consistency answers at the first epoch are based on those generated by Chinese-sampling without consistency at the first epoch. At the second epoch, the bilingual-consistency incorporates answers from both Chinese-sampling at the first epoch and English-sampling at the first epoch, and so on. In other words, during odd epochs of bilingual-consistency, the newly added answers come from Chinese-sampling, while during even epochs, the newly added answers come from English-sampling.

As the number of epochs increases, the accuracy of bilingual-consistency exhibits a two-stage pattern. In the first stage (from the 1st to the 7th epoch), the accuracy increases quickly from 78.0% to 87.0%, while in the second stage (from the 7th to the 20th epoch), the accuracy basically stabilizes, fluctuating between 86.7% and 87.5%. Take the 10th epoch as a comparison point, bilingual-consistency achieves an accuracy of 86.8%, demonstrating a significant accuracy improvement compared to self-consistency in Chinese (+1.0%) and self-consistency in English (+9.5%) on CMATH using DeepSeek-Math-7B-Instruct.

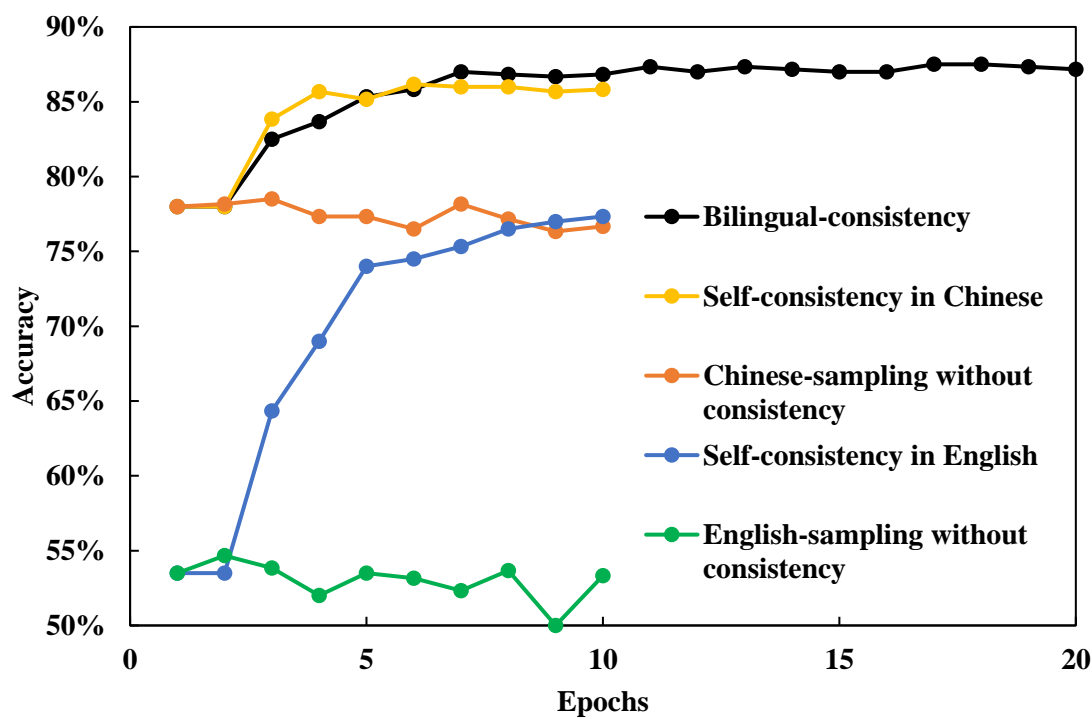

**Figure 2**. A comparison among bilingual-consistency, traditional self-consistency and sampling without consistency using DeepSeek-Math-7B-Instruct on CMATH.

The effectiveness of bilingual-consistency was validated on a broader range of LLMs and datasets. Figure 3 shows the accuracy results of bilingual-consistency, traditional self-consistency and sampling without consistency using Qwen2.5-Math-7B-Instruct on CMATH. In contrast to DeepSeek-Math-7B-Instruct, Qwen2.5-Math-7B-Instruct exhibits better reasoning ability on CMATH in English than in Chinese. When using Chinese-sampling without consistency, Qwen2.5-Math-7B-Instruct scores 67.2%-71.3% over 10 epochs, whereas it achieves 84.8-86.7% over 10 epochs with English-sampling without consistency. Similar to DeepSeek-Math-7B-Instruct, Qwen2.5-Math-7B-Instruct can achieve higher accuracy when self-consistency is deployed, leading to 83.8% and 87.5% accuracy at the 10th epoch in Chinese and English, respectively. More importantly, the effectiveness of self-consistency in English is quite weak with only +0.8% improvement compared to English-sampling without consistency. Therefore, it is quite impressive that bilingual-consistency can further enhance the reasoning ability of Qwen2.5-Math-7B-Instruct when self-consistency in English appears to have hit its performance bottleneck. As shown in Figure 3, the accuracy of bilingual-consistency increases quickly from 69.7% to 87.3% over the first four epochs, due to the distinct performance gap between the Chinese reasoning ability and English reasoning ability of Qwen2.5-Math-7B-Instruct. Then bilingual-consistency exhibits a mild increasing trend from the 4th to the 10th epoch with accuracy increasing from 87.3% to 90.3%, which is mainly attributed to the

Chinese reasoning ability of Qwen2.5-Math-7B-Instruct. Finally, the accuracy of bilingual-consistency stabilizes at the range of 89.8%-90.5%, achieving a high accuracy of 90.2% at the 20th epoch. The results demonstrate that bilingual-consistency significantly improves the reasoning performance of Qwen2.5-Math-7B-Instruct compared to self-consistency in Chinese (+6.5%) and self-consistency in English (+2.8%) on CMATH.

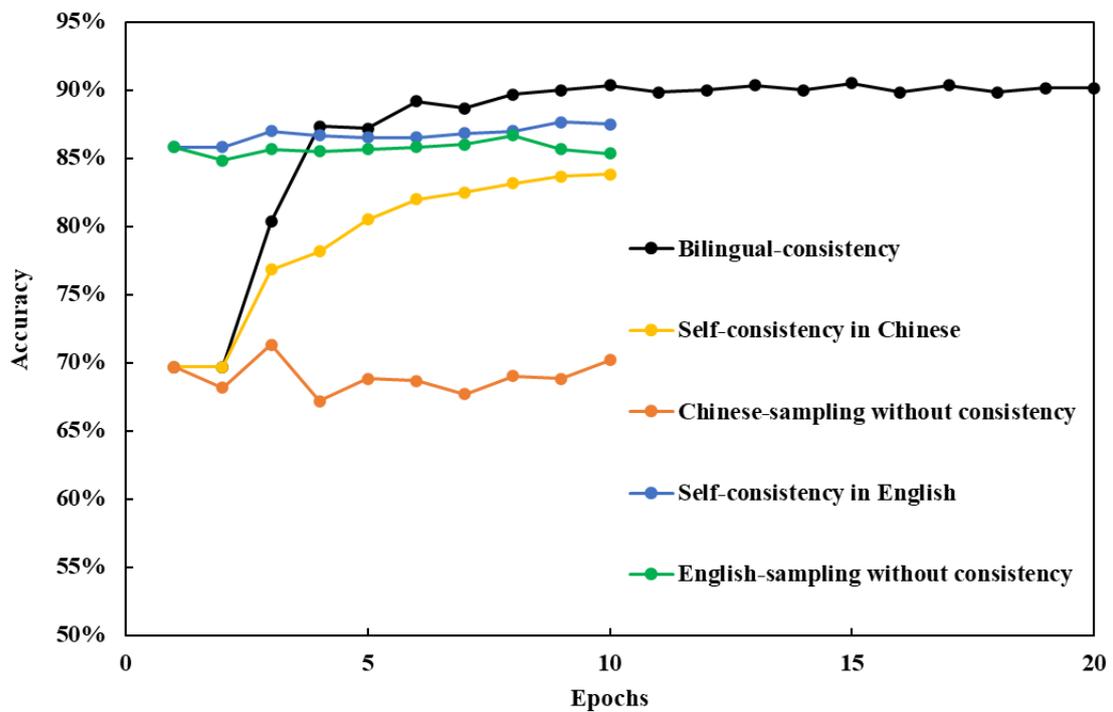

**Figure 3**. A comparison among bilingual-consistency, traditional self-consistency and sampling without consistency using Qwen2.5-Math-7B-Instruct on CMATH.

Similarly, the effectiveness of bilingual-consistency is also verified on Gemma2-9B-Instruct. As shown in Figure 4, the average accuracy difference of Gemma2-9B-Instruct between Chinese-sampling (average accuracy: 79.5%) and English-sampling (average accuracy: 75.6%) is about 3.9% on CMATH, significantly narrower than that of DeepSeek-Math-7B-Instruct or Qwen2.5-Math-7B-Instruct. In this case, Bilingual-consistency still exhibits significant accuracy enhancement compared with traditional self-consistency. Bilingual-consistency of Qwen2.5-Math-7B-Instruct scores 83.5% accuracy at the 10th epoch on CMATH, achieving +0.8% absolute gain over self-consistency in Chinese (82.7%) and +6.0% absolute gain over self-consistency in English (77.5%).

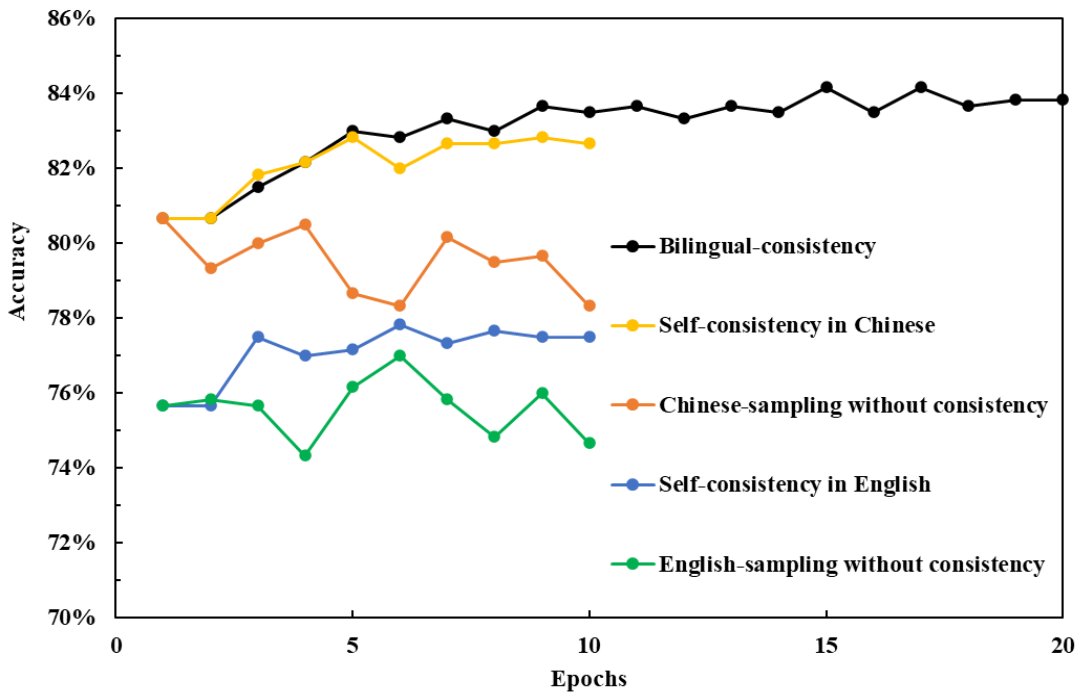

**Figure 4**. A comparison among bilingual-consistency, traditional self-consistency and sampling without consistency using Gemma2-9B-Instruct on CMATH.

It is intuitive that the smaller the performance gap between Chinese and English is, the less effective bilingual-consistency will be. However, the results of Gemma2-9B-Instruct on MGSM show that this is not the case. As shown in Figure 5, Chinese-sampling without consistency and English-sampling without consistency exhibit similar reasoning performance on MGSM using Gemma2-9B-Instruct. Specifically, the average accuracy of Chinese-sampling without consistency is 69.4%, while the average accuracy of English-sampling without consistency is 69.5%. As expected, self-consistency can increase the accuracy compared with sampling. Take the 10th epoch as a comparison point, self-consistency in Chinese achieves an accuracy of 75.2% which is +4.0% higher than that of Chinese-sampling without consistency. Self-consistency in English scores 71.6% which is +2.8% higher than English-sampling without consistency. Moreover, it is surprising that bilingual-consistency is still effective given that Chinese-sampling and English-sampling without consistency score almost the same. Bilingual-consistency achieves a accuracy of 77.6% at the 10th epoch, which is +2.4% higher than self-consistency in Chinese (75.2%) and +6.0% higher than self-consistency in English (71.6%).

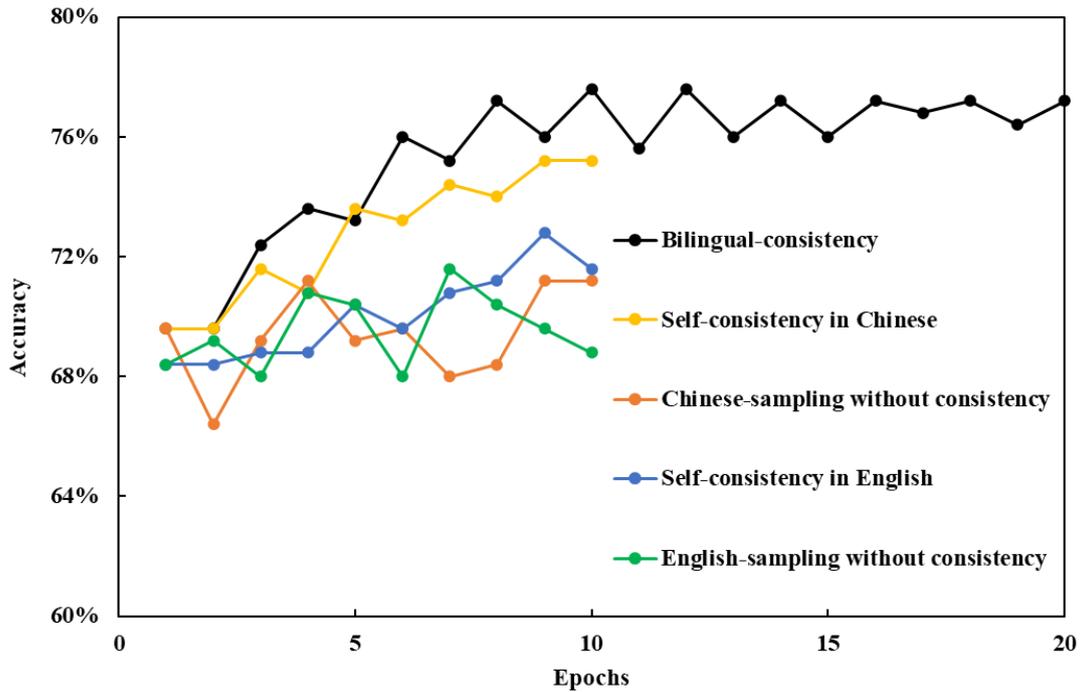

**Figure 5**. A comparison among bilingual-consistency, traditional self-consistency and sampling without consistency using Gemma2-9B-Instruct on MGSM. (Note: The accuracy values in Figure 5 are lower than those in Figure 6, due to different answer extraction functions).

**3.2.2 Multilingual-Consistency**

The effectiveness of CLC was also verified under the condition of multiple languages. MGSM, a dataset written in 11 languages, is suitable for the multilingual-consistency task. Figure 6(a) shows the accuracy comparison between multilingual-consistency and traditional self-consistency in each individual language, including Chinese, English, Bengali, German, Spanish, French, Japanese, Russian, Swahili, Telugu, and Thai, on MGSM using Gemma2-9B-Instruct. Among the 11 languages, self-consistency in English achieves the highest accuracy of 88.0% at the 10th epoch, followed by self-consistency in Spanish (87.6%), Russian (86.8%), French (82.4%), German (82.4%), Thai (81.6%), Chinese (80.0%), Swahili (80.0%), Bengali (76.8%), Japanese (75.6%), and Telugu (73.6%). The observed accuracy disparity likely stems from data sparsity challenges in low-resource languages like Telugu and Bengali, where limited training corpora and insufficient multilingual alignment hinder effective learning, whereas high-resource languages (English/Spanish) benefit from richer linguistic data to establish good performance.

The language integration sequence for multilingual-consistency was randomly chosen, which is Chinese, English, Bengali, German, Spanish, French, Japanese, Russian, Swahili, Telugu, and Thai. Therefore, the accuracy of Multilingual-consistency at the 1st epoch was 80.0%, the same as that of self-consistency in Chinese at the 1st epoch. Moreover, multilingual-consistency integrating 11 languages achieves 90.8% at the 11th epoch, which is 2.8% higher than self-consistency in English at the 10th epoch.

As shown in Figure 6(b), An accuracy of 91.6% is achieved by multilingual-consistency at the 110th epoch where data from 110 epochs (11 languages, 10 epochs for each language) were all used for majority voting. Interestingly, multilingual-consistency achieves its best accuracy, that is 92.0%, at the 24th epoch, which occurs at the early stage of the overall 110 epochs. However, the accuracy values of multilingual-consistency from the 25th epoch to 110th epoch range from 90.4% to 91.6%, never reaching 92.0% again. This implies that the performance of multilingual-consistency does not always increase as more reasoning languages are added. In other words, some reasoning languages have negative effects on the overall performance of multilingual-consistency.

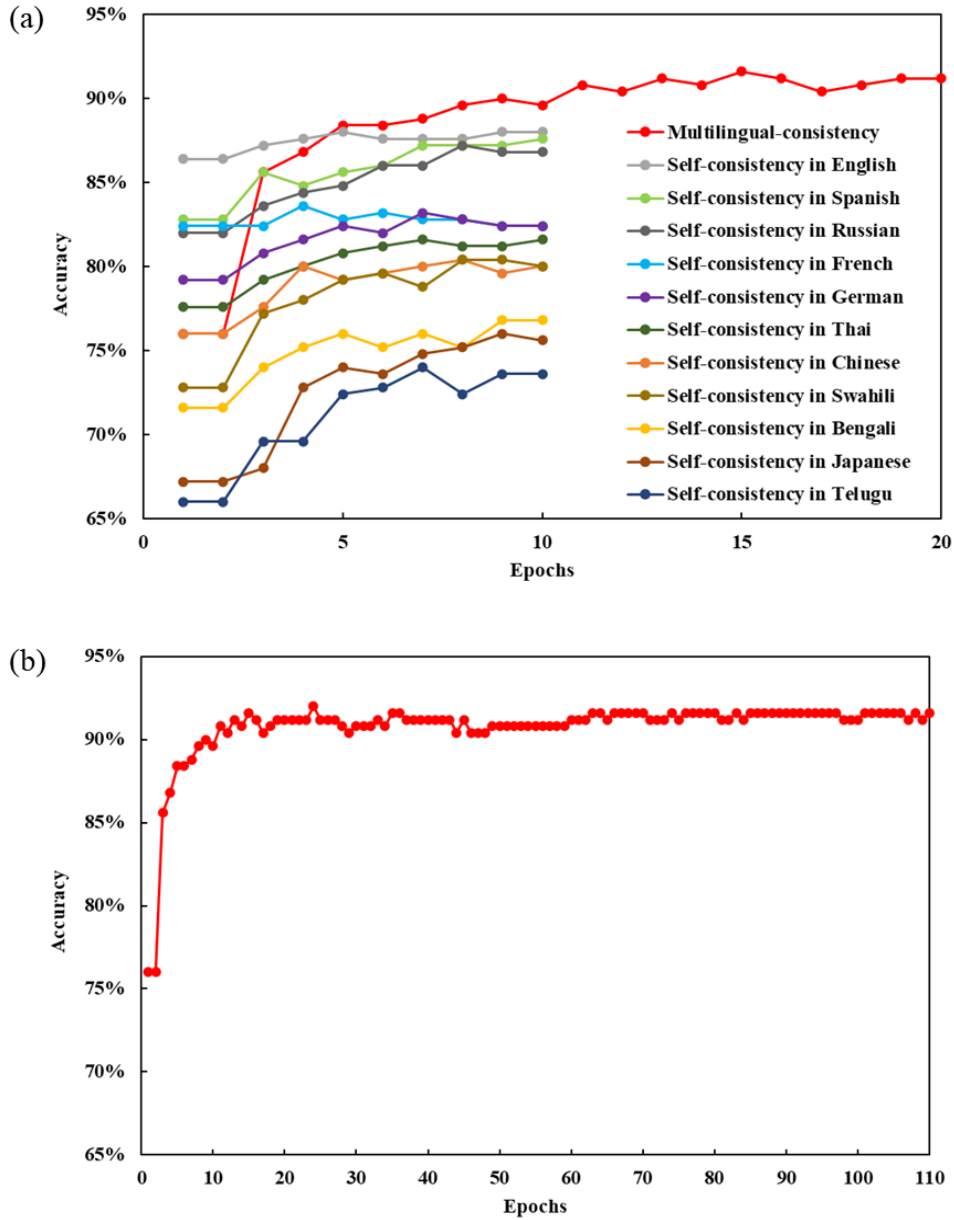

**Figure 6**. (a): A comparison between multilingual-consistency and traditional self-consistency in each individual language, including Chinese, English, Bengali, German, Spanish, French, Japanese, Russian, Swahili, Telugu, and Thai, using Gemma2-9B-Instruct on MGSM. (b): Accuracy of multilingual-consistency using Gemma2-9B-

Instruct on MGSM.

**4 Discussion**

It remains unclear how to predict and decide the optimal set of languages to achieve the best performance. Given the limited scale of the MGSM benchmark (250 samples), an exhaustive enumeration approach can be employed to compute accuracy metrics across all possible linguistic configurations. This methodology facilitates systematic identification of optimal language combinations for enhancing mathematical reasoning capabilities in large language models (LLMs). To ensure robust performance evaluation, we adopt the mean accuracy across post-20th-epoch iterations as the primary metric. This design intentionally mitigates two confounding factors: (1) convergence-induced accuracy instability during pre-20th-epoch iterations, and (2) post-convergence accuracy oscillation during post-20th-epoch iterations that may distort model capability assessment.

Given the combinatorial explosion of language configurations that generates 2,047 potential combinations (derived from $\sum_{n=2}^{11} C(11, n)$ ), Table 2 selectively presents the top-performing configurations within each group size (*n*=2 to 11) based on the mean accuracy metric across post-20th-epoch iterations. This hierarchical filtering strategy optimizes analytical tractability while preserving performance trends across combinatorial scales, effectively mitigating information redundancy inherent in exhaustive enumeration.

| combinatorial scales | Language combinations | Mean accuracy |
|---|---|---|
| $n=2$ | (English, Russian) | 90.80% |
| $n=3$ | (Spanish, Russian, Swahili) | 91.35% |
| $n=4$ | (Chinese, English, Russian, Swahili) | 91.87% |
| $n=5$ | (Chinese, English, Spanish, Russian, Thai) | 92.03% |
| $n=6$ | (Chinese, English, Bengali, Spanish, Russian, Thai) | 92.09% |
| $n=7$ | (Chinese, English, Bengali, German, Spanish, Russian, Thai) | 92.08% |
| $n=8$ | (Chinese, English, Bengali, German, Spanish, Russian, Swahili, Thai) | 91.94% |
| $n=9$ | (Chinese, English, Bengali, German, Spanish, Japanese, Russian, Swahili, Thai) | 91.90% |
| $n=10$ | (Chinese, English, Bengali, German, Spanish, French, Japanese, Russian, Swahili, Thai) | 91.50% |
| $n=11$ | (Chinese, English, Bengali, German, Spanish, French, Japanese, Russian, Swahili, Telugu, Thai) | 91.26% |

**Table 2**. Top-performing language configurations within each group size ($n=2$ to 11) based on the mean accuracy metric across post-20th-epoch iterations.

Through exhaustive enumeration of language configurations, we identify an optimal combination comprising six languages (Chinese, English, Bengali, Spanish,

Russian, and Thai), which achieves a mean accuracy of 92.09%. Therefore, compared to monolingual self-consistency with each language (73.6%-88.0%), CLC with the optimal combination achieves 4.1%–18.5% absolute accuracy gains on the MGSM benchmark using Gemma2-9B-Instruct, with particularly pronounced improvements in low-resource language scenarios.

As demonstrated in Table 2, the mean accuracy initially increases with the expansion of linguistic diversity, peaking at $n=6$, followed by a gradual decline. This phenomenon aligns with two competing mechanisms:

**Pattern coverage enhancement.** Multilingual ensemble voting mechanisms leverage parallel per-language inference pipelines to maximize coverage of training data patterns. Incorrect predictions from a single language can be overridden by correct responses from other languages through majority voting, effectively achieving pattern coverage via probabilistic consensus (Dietterich, 2000; Naik, 2024; Zhou, 2025). There are two synergistic benefits behind the probabilistic consensus of multilingual ensemble voting mechanisms: 1) neutralizing linguistic biases in multilingual training corpora through multilingual ensemble voting, 2) escaping monolingual reasoning traps by exploring the broader multilingual solution space.

**Probabilistic consensus degradation.** Beyond a critical cardinality ($n>6$), the system enters a phase of probabilistic consensus degradation (Dietterich, 2000; Naik, 2024; Zhou, 2025). Newly introduced languages may introduce conflicting predictions, overriding previously correct predictions through majority voting mechanisms.

**5 Conclusion**

While self-consistency mechanisms enhance CoT reasoning in large language models (LLMs), their efficacy is fundamentally constrained by multilingual training data biases—manifesting as semantic drift and logical inconsistencies, particularly in sub-10B parameter models. To overcome these limitations, we propose the CLC inference framework, which innovatively integrates multilingual CoT paths by majority voting to achieve performance gains. In the case of bilingual consistency, CLC achieves 9.5%, 6.5%, and 6.0% absolute accuracy gains on CMATH for DeepSeek-Math-7B-Instruct, Qwen2.5-Math-7B-Instruct, and Gemma2-9B-Instruct respectively, significantly outperforming traditional self-consistency baselines. Moreover, in the case of multilingual consistency, CLC achieves 4.1%–18.5% accuracy gains using Gemma2-9B-Instruct on the MGSM dataset compared to monolingual self-consistency with each language, with particularly pronounced improvements in low-resource language scenarios. There are two synergistic benefits behind CLC: 1) neutralizing linguistic biases in multilingual training corpora through multilingual ensemble voting, 2) escaping monolingual reasoning traps by exploring the broader multilingual solution space.

Future work should explore integration of CLC with retrieval-augmented generation, validate its efficacy in large-parameter models (>100B) like GPT-4 and PaLM-2, and extend the CLC inference framework to low-resourced language families.